%% file: MAIN_workshop.tex
\documentclass{article}

 \usepackage[dblblindworkshop, final]{neurips_2025}

\usepackage{enumitem}
\usepackage{comment}
\usepackage[utf8]{inputenc} % allow utf-8 input
\usepackage[T1]{fontenc}    % use 8-bit T1 fonts
\usepackage{hyperref}       % hyperlinks
\usepackage{url}            % simple URL typesetting
\usepackage{booktabs}       % professional-quality tables
\usepackage{tabularray} % Even better tables
\usepackage{amsmath}
\usepackage{amsfonts}       % blackboard math symbols
\usepackage{nicefrac}       % compact symbols for 1/2, etc.
\usepackage{microtype}      % microtypography
\usepackage{xcolor}         % colors
\usepackage{listings}
\usepackage{float} % in preamble
\usepackage{graphicx}
\usepackage{multirow}
\usepackage{bbm}

\newcommand{\redcomment}[1]{\textcolor{red}{[#1]}}

\title{WebATLAS: An LLM Agent with Experience-Driven Memory and Action Simulation}
\workshoptitle{LAW 2025: Bridging Language, Agent, and World Models for Reasoning and Planning}

\author{%
  Jiali Cheng\thanks{Work done during internship at Amazon} \\
  University of Massachusetts Lowell \\
  \AND
  Anjishnu Kumar \\
  Amazon Alexa AI \\
  \And
  Roshan Lal \\
  Amazon Alexa AI \\
  \And
  Rishi Rajasekaran \\
  Amazon Alexa AI \\
  \AND
  Hani Ramezani \\
  Amazon Alexa AI \\
  \And
  Omar Zia Khan \\
  Amazon Alexa AI \\
  \And
  Oleg Rokhlenko \\
  Amazon Alexa AI \\
  \AND
  Sunny Chiu-Webster \\
 Amazon Alexa AI \\
  \And
  Gang Hua \\
Amazon Alexa AI \\
  \And
  Hadi Amiri \\
  University of Massachusetts Lowell \\
}

\begin{document}

\maketitle

\input{000abstract}
\input{010intro}

\input{020related}
\input{030method}
\input{040experiment}

\input{050results}

\input{070conclusion}

\bibliographystyle{ICLR2026/iclr2026_conference}
\newpage
\bibliography{reference}

%%%%%%%%%%%%%%%%%%%%%%%%%%%%%%%%%%%%%%%%%%%%%%%%%%%%%%%%%%%%

%\newpage
%\appendix
%\input{090appendix}

%%%%%%%%%%%%%%%%%%%%%%%%%%%%%%%%%%%%%%%%%%%%%%%%%%%%%%%%%%%%

\end{document}

%% file: 000abstract.tex
\begin{abstract}

Large Language Model (LLM) web agents often struggle with long-horizon web navigation and web task completion in new websites, producing inefficient action sequences unless fine-tuned on environment-specific data. We show that \emph{experience-driven memory}, combined with \emph{look-ahead action simulation}, is sufficient for LLM agents to adapt to unseen web environments by remembering past failures and predicting the consequences of future actions. We introduce \textbf{WebATLAS} (\textbf{A}ctor-Critic \textbf{T}ask-completion with \textbf{L}ook-ahead \textbf{A}ction \textbf{S}imulation), a memory-augmented LLM web agent that learns a lightweight internal model of the environment from interaction experience and performs hypothetical action rollouts before acting in the real world. WebATLAS builds a persistent \emph{cognitive map} via curiosity-driven exploration, stores interaction outcomes as experience-based memory, and evaluates candidate actions in cognitive space using a planner--simulator--critic loop. This enables the agent to reuse past experience, avoid previously unsuccessful behaviors, and generate more efficient plans. We evaluate WebATLAS on the \textbf{WebArena-Lite} benchmark for autonomous web navigation and demonstrate a success rate of 63\%, outperforming the previous state-of-the-art at 53.9\%. Unlike previous systems, our modular architecture requires no website-specific LLM fine-tuning. Ablation studies confirm that experience-driven memory, look-ahead action simulation, and hierarchical replanning play complementary roles in enabling robust, training-free web agents.

\end{abstract}

%% file: 010intro.tex
\section{Introduction}

Autonomous agents that can navigate and act on the web have the potential to perform complex tasks like information gathering, transactions, and site configuration on behalf of users \citep{yao2022webshop, step}. However, current web-based agents fall far short of human-level reliability on long-horizon tasks \citep{koh2024tree}. The difficulty stems from partial observability, vast action spaces, the need for multi-step planning and memory in a web environment. For example, tasks in WebArena~\citep{webarena} span diverse websites and require reasoning over multi-page navigation and content. An agent might be asked to “Tell me how many fulfilled orders I have over the past three days, and the total amount spent” on an e-commerce site, or “Set the homepage URL on my GitLab profile to https://egg.tart.com.” Solving these tasks demands understanding site structure (e.g. admin dashboards, forms), remembering relevant information (like login states or filters applied), and avoiding irreversible mistakes (like deleting data or purchasing an item unintentionally).

Large language models (LLMs) have markedly improved semantic understanding and generation, suggesting they can enrich web navigation. Yet reliable long-horizon control remains elusive because LLM agents are typically reactive and lack structured memory and explicit planning. Current state-of-the-art web agents such as Plan-and-Act by \cite{erdogan2025planandact} have significant gaps - the agent's planner module is not grounded to the structure of the website and requires website-specific model fine-tuning of both planner and executor modules in order to enable new use-cases.

We address these gaps with WebATLAS—an inference-time, actor–critic web agent that plans before acting via look-ahead simulation and retrieves structured memories to remain goal-directed over extended interactions. WebATLAS has a modular architecture with four components: (1) a Planner that decomposes the task into subgoals; (2) an Actor that proposes diverse next-step candidates; (3) a Critic that forecasts each candidate’s outcome by simulating state transitions and selects the safest, goal-advancing action; and (4) a multi-layer memory that is updated online and queried on demand. Together, these modules perform a simulated look-ahead tree search in conceptual space, enabling adaptive environment-grounded planning, and efficient action selection in realistic, partially observable websites, including those with unexpected \emph{environmental hazards}.

Our agent uses a hierarchical memory structure: (i) Working Memory for recent context; (ii) a Cognitive Map encoding state transitions and expected action outcomes; and (iii) Semantic Memory that captures environment-specific constraints (e.g. formats, hazards) in the form of world knowledge. The map is constructed via curiosity-driven exploration (MCTS-style trajectory mining) and agentic summarization that records action-to-outcome deltas in natural language, avoiding HTML bloat. During inference, the Actor conditions on the plan, retrieved memories, and in-context trajectories; the Critic filters out risky or myopic moves by simulating their consequences.

We summarize our contributions as the following:
\begin{itemize}[leftmargin=*, labelsep=0.75em]
    \item An \emph{actor-critic planner} with LLM-based look-ahead that evaluates actions via simulated outcomes.
    \item A \emph{multi-layer memory} with a cognitive map built through exploration and agentic summarization, used online for retrieval and replanning. 
    \item A \emph{practical modular architecture} that integrates \emph{planning}, \emph{memory}, and \emph{simulation} to transform high-level instructions into safe, executable action sequences for long-horizon web tasks. Unlike previous systems, our system does not require website-specific LLM fine-tuning to ground to new websites and can thus easily be ported to new websites and new underlying LLMs.
\end{itemize}

%% file: 020related.tex
\section{Related Work}

\paragraph{LLM-Based Autonomous Agents}

The integration of large language models into autonomous agents has revolutionized web navigation capabilities. \cite{yao2022react} pioneered the ReAct framework, demonstrating how LLMs can effectively combine reasoning and acting in interactive environments. This approach has been extended by \cite{shinn2024reflexion} with Reflexion, which incorporates self-reflection mechanisms for improved decision-making over extended horizons.

\paragraph{Web Navigation Agents}

Early web navigation systems relied predominantly on rule-based approaches and predefined scripts \cite{liu2018reinforcement}, which, while interpretable, lacked the adaptability required for dynamic web environments. Recent advances have shifted toward learning-based methods, with several notable developments in autonomous web agents. \cite{webarena} introduced WebArena, a comprehensive benchmark for evaluating web agents across realistic multi-step tasks, establishing a foundation for systematic evaluation in this domain. Building on this, \cite{liu2024vab} introduced WebArena-Lite, a curated subset addressing quality and scalability concerns in the original benchmark. 

Recent work has focused on enhancing LLM agents with structured approaches to planning and memory and demonstrated significant progress: \cite{qi2024webrl} applied reinforcement learning principles to web navigation, achieving notable improvements through policy optimization. \cite{zhang2025webpilot} developed WebPilot, focusing on multimodal understanding of web content, while \cite{erdogan2025planandact} introduced Plan-and-Act, emphasizing hierarchical task decomposition. \cite{wang2024agentworkflowmemory} introduced Agent Workflow Memory (AWM), demonstrating the importance of persistent memory in multi-step web tasks. \cite{yang2024agentoccam} developed AgentOccam and demonstrated the effectiveness of simplifying the action space to natural language.

\paragraph{Memory-Augmented Agents}

Memory mechanisms have emerged as crucial components for long-horizon task completion. \cite{weston2014memory} established foundational work on memory networks, which has been adapted for sequential decision-making contexts. More recently, \cite{zhong2024memorybank} proposed MemoryBank, a comprehensive framework for managing episodic and semantic memory in LLM-based agents. The concept of cognitive maps, originally from cognitive science \cite{tolman1948cognitive}, has been adapted for artificial agents. \cite{wayne2018unsupervised} demonstrated neural implementations of cognitive mapping in reinforcement learning contexts, while  \cite{park2023generative} showed how LLMs can maintain and utilize spatial-temporal memory representations for complex behavioral simulation.

\paragraph{Planning and Simulation in AI Agents}

Tree search and simulation-based planning have long been central to AI agent design. While classical approaches like Monte Carlo Tree Search \cite{browne2012survey} have proven effective in discrete domains, recent work has extended these concepts to natural language environments. \cite{yao2024tree} introduced Tree of Thoughts, enabling LLMs to explore multiple reasoning paths through structured search.

World models, which enable agents to simulate future states without environment interaction, have gained renewed attention. \cite{ha2018world} established foundational work on learned world models, while more recent efforts by \cite{micheli2022transformers} have shown how transformer architectures can serve as effective world models for sequential decision-making.

\paragraph{Actor-Critic Methods and Look-ahead Planning}

Actor-critic architectures have proven particularly effective for combining policy learning with value estimation. While traditional actor-critic methods focus on reinforcement learning training \cite{sutton2018reinforcement}, recent work has adapted these principles to LLM-based agents. The integration of look-ahead planning with actor-critic frameworks has been explored in various contexts, with \cite{silver2016mastering} demonstrating the power of combining tree search with learned value functions in AlphaGo.

\paragraph{Curiosity-Driven Exploration}

Exploration remains a fundamental challenge in autonomous agent design. \cite{pathak2017curiosity} introduced intrinsic curiosity modules that drive exploration through prediction error, establishing a foundation for self-supervised exploration. Recent work has extended these concepts to language-based environments: \cite{mu2024embodiedgpt} demonstrated curiosity-driven exploration in embodied AI settings.

Our work builds upon these foundations by combining memory-augmented planning with look-ahead simulation in a modular architecture specifically designed for web navigation tasks. Unlike previous approaches that require environment-specific fine-tuning, WebATLAS achieves strong performance through inference-time planning and memory retrieval, making it readily adaptable to new domains and underlying models.

%% file: 030method.tex
\section{Method}
\label{sec:method}

% Requires: \usepackage{amsmath,amssymb}

\subsection{Problem formulation}
We cast web navigation as a Partially Observable Markov Decision Process (POMDP) defined by the tuple $(\mathcal{S},\mathcal{A},\mathcal{O},T,R)$, where 
$\mathcal{S}$ denotes the state set, 
$\mathcal{A}$ denotes the action set,
$\mathcal{O}$ denotes the observation set,
$T$ denotes the state transition function, and
$R$ denotes the reward.
Given a natural-language goal $q$, the agent must synthesize a plan and execute a sequence of actions $(a_0,\ldots,a_T)$ that reaches a goal-consistent terminal state.
At each time step $t$, the agent receives partial observations $o_t \in \mathcal{O}$. Based on the observation $o_t$, the agent chooses an action to take $a_t \in \mathcal{A}$, such as \texttt{click} or \texttt{type}. The goal of the agent is to maximize the reward, i.e. fulfilling task $q$.
% latent states $s_t$, actions $a_t \in \mathcal{A}$ (click, type, select, stop), trasnsition kernel $\mathcal{T}$, and sparse task reward $\mathcal{R}$ (success/fail). 

% When $a_t = \texttt{stop}$, the evaluator outputs a binary reward (success or fail).

\subsection{Architecture Overview}
\begin{figure}[t]
    \centering
    \includegraphics[width=0.99\linewidth]{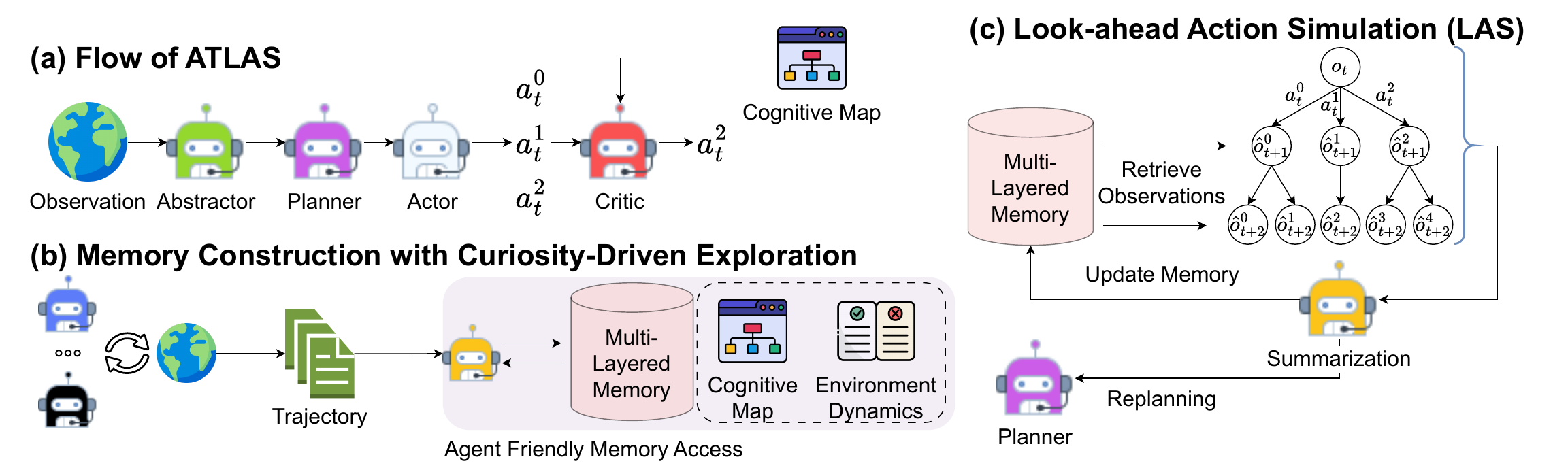}
    \caption{\textit{\textbf{Architecture of WebATLAS.}
    (a) \textbf{Overall flow of WebATLAS:} The raw observation $o_t$ is summarized to lower cognitive load. Then the planner makes a plan $P_t$ based on the summarized observation $o'_t$. The actor proposes $N$ possible candidate actions for next step. The critic provides judgment of action candidates and finalizes the best action $a_t$ to take by considering action outcomes obtained from the cognitive map. 
    (b) \textbf{Memory construction with curiosity-driven exploration:} We build cognitive map by employing exploratory lightweight agents to interact with the environment. 
    (c) \textbf{Look-ahead Action Simulation (LAS):} At each step, \textbf{WebATLAS} simulates all candidate actions with the observation from the cognitive map, providing ability to look-ahead. We employ the memory agent to learn from LAS trajectories to make a better plan and update memory if necessary.}}
    \label{fig:atlas}
\end{figure}

WebATLAS comprises four modules operating in an inference-time \emph{actor--critic} loop with \emph{action simulation in conceptual space}, shown in Figure~\ref{fig:atlas} (a):

\begin{enumerate}[label = (\arabic*)]%[leftmargin=*, labelsep=0.5em, nosep]
  \item \textbf{Planner} produces a high-level plan with subtasks, with the ability to replan;
  \item \textbf{Actor} proposes a small set of next-step candidates $C_t$; % $(q,P,o_t,s_t,M)$;
  \item \textbf{Critic} performs outcome-aware simulation of each candidate and selects the safest, goal-advancing action;
  \item \textbf{Multi-layered memory} supplies working context, a cognitive map of state transitions, and world knowledge about the environment; it is queried online and updated as needed.
\end{enumerate}

\paragraph{Planner}
The planner analyzes and decomposes the natural language task $q$ into a structured plan with subtasks to finish. Given the initial observation $o_0$, the planner produces an initial plan $P_0$; at step $t$, it dynamically decides if the plan needs to be updated (replanning) given new evidence
\vspace{-0.5em}

\begin{equation}
  P_0=\mathrm{Planner}(q,o_0), \qquad
  P_t=\mathrm{Planner}(q,o_t,s_t,M).
\end{equation}

Plans are concise lists of sub-goals with success predicates (e.g., ``Reports $\rightarrow$ Sales $\rightarrow$ Set dates $\rightarrow$ Read table''). Planner outputs are included in the context for the actor and critic. Our planner is implemented in the style of \citep{webshepherd} and extended and described in \ref{sec:replanning}.

\paragraph{Actor-Critic interplay with look-ahead}\label{sec:actor_critic}
In our framework, at each step $t$, the actor proposes $N$ executable candidates with reasoning, and the critic evaluates it based on a value function $V(a)$. 

\vspace{-0.5em}
\begin{equation}\label{eq:actor}
  C_t=\mathrm{Actor}(q,P_t,o_t,s_t,M), \qquad |C_t|=N.
\end{equation}
\vspace{-0.5em}

The critic evaluates each candidate action $a^i_t \in C_t$ and selects the best next action
\begin{equation}
  a_t=\arg\max_{a\in C_t} V\!\left(a \mid q,P_t,o_t,s_t,M\right).
\end{equation}
\vspace{-0.5em}

The utility estimate $V(a)$ is derived via LLM-based assessment that incorporates goal alignment, state viability (recoverability), action coherence, plan consistency, and outcome risk (e.g., destructive or dead-end transitions). Unlike previous efforts that attempt to learn implicit world-model of the environment by fine-tuning a neural network model, such as \cite{wma}, we leverage the cognitive map to retrieve action outcomes of each action candidate (Section~\ref{sec:cog_map}). This gives the agent system the ability to look ahead of the current step. We later extend this standard behavior with simulated tree search to further enhance exploratory ability in Section~\ref{sec:tree_search}.

\paragraph{Multi-layered memory}
\label{sec:memory}
We use three complementary memories:
\begin{itemize}[leftmargin=*, labelsep=0.8em]
  \item \textbf{Working Memory}: A task-specific memory wherein facts and observations are optionally stored into the LLM context for use during a particular episode.
  \item \textbf{Cognitive Map}: A graph of transitions $M=\{(o,a,o')\}$ with \emph{agentic summaries} that store deltas and new affordances (e.g., ``click \texttt{Reports} reveals \{Sales, Products,\dots\}'') rather than raw HTML. The map supports retrieval $\hat o_{t+1}=M(o_t,a)$ for simulation and planning (Section~\ref{sec:cog_map}).
  \item \textbf{Semantic Memory (World Knowledge)}: Learned environment dynamics (e.g., date/format rules, non-recoverable states), used to penalize risky actions and inform simulation (Section~\ref{sec:env_dynamics}).
\end{itemize}

\newpage
\subsection{Memory construction via Curiosity-Driven Exploration}
\label{sec:exploration}

\paragraph{Motivation}
Existing agents may fail because they are 1) not aware of the potential action outcomes, such as placing an order on a shopping website likely to lead to difficulty of canceling and refunding, and, 2) not familiar with environment-specific requirements, such as date format and search format (e.g. AgentOccam-Judge on WebArena-Lite Task 0)~\citep{yang2024agentoccam}. This is a major gap between LLM agents and human intelligence, since humans can easily predict the outcome of an action with our world knowledge. Encouraging agents to explore the environment and store the findings in memory can effectively avoid actions leading to undesirable outcomes.

\paragraph{Memory construction}
Prior work on artificial curiosity \cite{pathak2017curiosity} has demonstrated the ability of neural networks operating in an agentic context to frame curiosity as a self-supervised learning task. Inspired by this insight, we augmented our agent with an artificial curiosity module to initialize the agent. Before evaluation, we perform a curiosity-driven exploration of the web environment to seed the cognitive map and world knowledge of the agent. 

The memory construction process consists of the following steps:
\begin{itemize}[leftmargin=*, labelsep=0.5em]
  \item \textbf{Exploration policies}: We first launch a series of lightweight explorer subagents with diverse LLM generation temperatures and exploration policies. We embed coverage incentives into the prompt of the explorer agents. No task-completion reward is used, in order to avoid information leakage from the test set. We balance breadth, depth, and entropy, limiting the explorers to visit the most promising states within a fixed memory budget.
  
  \item \textbf{LLM based trajectory-mining}: Given the exploration trajectories, we employ an LLM to convert the trajectories to agentic summaries of the environmental transitions, and store it as a cognitive map. Additionally, we employ an LLM to produce agentic summaries of site-specific rules, constraints, and hazards in semantic memory.
  
  % \item \textbf{MCTS/beam-style rollouts}: balance breadth, depth and entropy and limit the exploration to explore the most promising states within a fixed memory budget.
  % \item \textbf{LLM based trajectory-mining}: convert traces to transition triples and agentic summaries and store it in the cognitive map; record site-specific rules/constraints/hazards in semantic memory.
\end{itemize}

\paragraph{Memory layer 1: Cognitive Map}\label{sec:cog_map}
% The Cognitive map component encodes structural and navigational knowledge about the operational environment. 
% This is essentially a learned world model or transition model in RL, or Cognitive map about environmental facts.

The cognitive map encodes structured knowledge about the environment’s dynamics, including state transitions and causal relationships. Conceptually, it is akin to a learned world model or transition model in reinforcement learning, capturing how actions alter observations. For example, “clicking \texttt{Add to Cart} on a product page” results in a cart update notification, while “entering text in the search bar” leads to a results page. Formally, the cognitive map is represented by tuples $(o_t, a_t, o_{t+1})$, where $o_t$ and $o_{t+1}$ denote observations (e.g., HTML content or URLs in text-based web environments) at steps $t$ and $t+1$, and $a_t$ is the action executed at step $t$.  

At each step of exploration, we document the current observation $o_t$, the executed action $a_t$, and the subsequent observation $o_{t+1}$. To enhance interpretability and lower cognitive load of the agent, we adopt an \emph{agentic memory} strategy, where an LLM agent curates what is written into memory. Specifically, the memory agent produces concise summaries emphasizing (in addition to the raw observations):
\begin{itemize}
    \item Differences between successive observations $(o_t, o_{t+1})$;  
    \item Newly available actions in $o_{t+1}$ after executing $a_t$.  
\end{itemize}
% Note that the raw observations are also preserved to support downstream processes such as action simulation (Section~\ref{sec:tree_search}).  

For retrieval, the cognitive map is queried with $(o_t, a_t)$, returning the next raw observation $o_{t+1}$, as well as LLM summaries. This design balances fidelity (retaining raw states) with abstraction (summarized transitions), enabling efficient reasoning over complex, text-based environments. When the retrieval hits an unexplored node in the cognitive map, a generic-placeholder observation is returned. 
% This placeholder response signals a retrieval miss while preserving the structural integrity of the reasoning process. 
% (i) \emph{Cognitive Map-based retrieval} when the transition $(o_t,a)$ is known and (ii) an \emph{LLM hypothesis} otherwise, producing a next-state prediction $\hat o_{t+1}$ and a utility estimate $V(a)$. It then selects

\paragraph{Memory layer 2: Semantic Memory (World Knowledge)}\label{sec:env_dynamics}

This memory captures environment-specific knowledge, such as constraints, formats, and idiosyncratic behaviors unique to each website. For instance, it encodes facts like ``the date picker only accepts input in \texttt{MM/DD/YYYY} format'' or ``the admin portal does not support exporting tables into CSV files''. By recording these particulars from prior explorations, semantic memory serves as a bridge between specific past experiences and working memory, which maintains awareness of the immediate environmental context. This integration enables agents to adapt more effectively to recurring interface patterns and site-specific limitations.  The cognitive map and semantic memory are also optionally updated online when execution encounters unseen transitions or world dynamics.

% Note : for each of these memories we use \textit{Agentic summarization.} An LLM-based memory writer records only (i) differences $\Delta(o,o')$ and (ii) newly available actions after $a$, preventing HTML bloat and improving precision. This agentic summarization is a critical for our system to work.

\subsection{Look-ahead Action Simulation (LAS)}
\label{sec:tree_search}
The standard actor-critic interplay (Section~\ref{sec:actor_critic}) is a good baseline but may suffer from insufficient exploration and lack of foresight. To mitigate this issue, we propose Look-ahead Action Simulation (LAS). 

At step $t$, the actor first generates the action candidates $C_t$ as described in Equation~\ref{eq:actor}. For each action candidate $a^i_t \in C_t$, the critic hypothetically selects $a^i_t$ as the final action to execute and provides criticism. The resulting change of observation is retrieved from the cognitive map
\begin{equation}
    \hat{o}^i_{t+1} = M(o_t, a^i_t),
\end{equation} where $i$ is the index of the action candidate at step $t$.

We repeat this for $D$ times, resulting in a set of rolled-out trajectories of length $D$. 
% for top candidates it may extend depth to $D=2$ to avoid myopic choices. 
Let $\hat{\tau}$ denote a simulated trajectory (length $D$) with value $V(\hat \tau)$. We apply confidence weighting based on transition uncertainty $U(s,a)$:
\begin{equation}
  \tilde V(\hat{\tau})=V(\hat{\tau})\cdot\prod_{(s,a)\in\hat{\tau}}\bigl(1-U(s,a)\bigr).
\end{equation}
The best trajectory determines the real action $a_t$. 

\paragraph{Comparison to prior work} Existing agents perform tree search with LLMs as a reward function or a world model and measures the quality of each possible action candidate with a numerical score. Then the agent only executes actions whose scores are above a certain threshold. Our simulated tree search has three advantages compared to prior methods: 

\begin{enumerate}
    \item \emph{Trustworthiness}: Prior works rely on LLM to envision outcomes of actions. Since LLMs are not explicitly trained to be a world model, this behavior is prone to hallucination and is not robust enough. In contrast, our method leverages real observations that are much more trustworthy.
     \item \emph{Comprehensiveness}: Prior works essentially conducts a greedy search (one-step), with low-scoring branches directly pruned without further consideration. Some actions may not be good at the immediate step $t$ but useful at the next step $t+1$. Such actions may be overlooked in existing agent systems. While our method is similar to beam search (multi-step), considering the joint outcome of a sequence of actions.
    \item \emph{Efficiency}: Our exploration is a simulation in conceptual space, which is much more efficient than actually executing the actions. It also avoids stateful actions that cannot be recovered, since no action is executed.

\end{enumerate}
% This limited-depth beam yields most benefits of lookahead search without the overheads of costly environment branching.

% This loop realizes a shallow tree search in conceptual space: allowing us to simulate multiple trajectories in conceptual space before acting, execute the best branch as selected by the critic agent, observe the results, and re-plan and update our memory states when necessary. Our agent builds on the baseline Re-Act style agent introduced by \cite{yang2024agentoccam} who introduced a set of state space simplifications that made the task easier for an LLM based agent for perform. 

\subsection{Look-ahead Action Simulation-backed Dynamic Replanning and Memory Update}
\label{sec:replanning}

\paragraph{Replanning}
We dynamically trigger replanning when observations diverge from expectations:
\begin{equation}
  \mathrm{replan} = \mathbbm{1} \left[\ \lVert o^{\mathrm{obs}}_{t}-\hat o^{\mathrm{exp}}_{t}\rVert>\varepsilon \right].
\end{equation}

A task-relevant plan requires a high-level view of the environment and the ability to foresee what would happen in future steps. We attempt to distill the foresight enabled by simulated tree search (Section~\ref{sec:tree_search}), by using the result of the search to update our planner.

As Figure~\ref{fig:atlas} (c) illustrates, the planner integrates a brief \emph{exploration digest} (what worked/failed, newly exposed affordances, uncovered prerequisites) distilled by the memory writer, then updates $P_t$. This can be viewed as a highly simplified implementation of a basic causal learning  module that attempts to update our causal model of the world - a highly simplified version of the conceptual flow introduced by \cite{causalcuriosity}. This mechanism also prevents catastrophic forgetting of important context, which can happen if the replanner is run with every execution step.

\paragraph{Memory Update} In addition to replanning, the agent must also be capable of updating its memory during action simulation. This process applies to both the cognitive map and episodic memory, ensuring that newly encountered patterns, constraints, or dynamics are incorporated into long-term knowledge. Crucially, decisions about what to retain, update, or forget are delegated to the memory agent, which curates information based on task relevance and environmental novelty. Such selective updating is particularly important during curiosity-driven exploration, where novel experiences can refine the agent’s representation of the environment while preventing memory overload with redundant or irrelevant details.  

%% file: 040experiment.tex
\section{Experimental Setup}

WebArena~\citep{webarena} presents a realistic simulation environment comprising a broad array of web navigation tasks such as content retrieval, task execution, and form completion. Tasks vary in complexity, ensuring that the agent’s capabilities are thoroughly tested in realistic scenarios - ranging from purchasing items for eCommerce shopping to updating GitLab code repositories.

The original WebArena consists of 811 tasks, however many of these cannot be performed - humans could only perform 78\% of the WebArena. WebArena-Lite is a quality-controlled smaller subset of WebArena consisting of 165 tasks introduced by~\citep{liu2024visualagentbench} and has been adapted by prior work in the web agent space such as WebRL~\citep{qi2024webrl} and Plan-and-Act~\citep{erdogan2025planandact} as a higher quality and more scalable benchmark for evaluating web agents in the most realistic setting possible - incorporating realistic scenarios such as unexpected environment failures.

%% file: 050results.tex
\section{Results}

\subsection{Comparison with Other Baselines}

In this section, we compare our model to other published results on the WebArena-Lite dataset. These are WebPilot~\citep{zhang2025webpilot}, Agent Workflow Memory (AWM)~\citep{wang2024agent}, WebRL \cite{qi2024webrl}, and Plan-and-Act~\citep{erdogan2025planandact}. Our own work builds on top of AgentOccam~\citep{yang2024agentoccam} - we rerun our results on AgentOccam using the state-of-the-art LLM available to us for experimentation at scale, Claude-4 Sonnet.

\begin{table}[t]
\caption{Evaluation Results for WebATLAS versus other methods reported on WebArena-Lite. Best performance is in \textbf{bold}}
\centering
\setlength{\tabcolsep}{2pt}
\begin{tabular}{lcccccccc}
\toprule
Agent & \begin{tabular}[c]{@{}c@{}}Avg w/\\Multi-site\end{tabular} & \begin{tabular}[c]{@{}c@{}}Avg w/o\\Multi-site\end{tabular} & Gitlab & Reddit & Shopping & \begin{tabular}[c]{@{}c@{}}Shopping\\Admin\end{tabular} & Maps & \begin{tabular}[c]{@{}c@{}}Multi-\\Site\end{tabular} \\
\midrule
WebPilot + GPT-4o & - & 35.3 & 39.4 & 65.1 & 36.9 & 24.7 & 33.9 & -- \\
\addlinespace

AWM + GPT-4-0613 & - & 33.0 & 31.8 & 50.9 & 30.8 & 29.1 & 43.3 & -- \\
\addlinespace

WebRL & - & 48.1 & 50.0 & 78.9 & 44.4 & 54.3 & 40.0 & -- \\
\addlinespace

Plan-and-Act & 53.9 & 57.5 & 53.3 & 84.2 & \textbf{55.6} & 48.6 & 46.6 & 30.0 \\
\addlinespace

\begin{tabular}[c]{@{}l@{}}AgentOccam \\ \textit{(Claude-4-Sonnet)}\end{tabular} & 47.9 & 51.0 & 66.7 & 63.2 & 40.0 & 54.3 & 23.1 & 40.0 \\
\addlinespace

%\begin{tabular}[c]{@{}l@{}}AGENTOCCAM-Judge \\ (Claude-4-Sonnet)\end{tabular} & 46.7 & 49.7 & 66.7 & 68.4 & 40.0 & 42.9 & 30.8 & 30.0 \\
\addlinespace
\midrule
\textbf{WebATLAS (Ours)} & \textbf{63.0} & \textbf{67.1} & \textbf{73.3} & \textbf{84.2} & 53.3 & \textbf{77.1} & \textbf{42.3} & \textbf{40.0} \\
\bottomrule
\end{tabular}
% \addlinespace
\end{table}

\subsection{Ablation Study}

In this section we conduct an ablation study where we study the effects of the different components of the system.  Starting with AgentOccam as our base agent, we demonstrate that augmenting the agent with the direct HTML cognitive map \textit{(Base + CM-Raw)} initially reduced performance but led to a dramatic improvement in performance after we enabled agentic summarization (Base + CM). 

In addition, integrating a high-level planner \textit{(Base + HL)} in the style of \cite{webshepherd} also improves performance on top of the base agent. 

Finally, we construct the final WebATLAS agent by integrating both the cognitive map and high-level planner and further extend the system to include replanning via look-ahead search \emph{(Base + CM + HL + LA)} in order to condition the planner on the cognitive map, we see that the two systems demonstrate complementary performance to produce superior performance in conjunction - achieving state of the art results on WebArena-Lite. 

\begin{table}[t]
\caption{Ablation Study Results for Individual Components of \textbf{WebATLAS}.}
\centering
\setlength{\tabcolsep}{2pt}
\begin{tabular}{lcccccccc}
\toprule
Agent & \begin{tabular}[c]{@{}c@{}}Avg w/\\ Multi-site\end{tabular} & \begin{tabular}[c]{@{}c@{}}Avg w/o\\ Multi-site\end{tabular} & Gitlab & Reddit & Shopping & \begin{tabular}{c}
     Shopping \\
     Admin
\end{tabular} & Maps & \begin{tabular}[c]{@{}c@{}}Multi-\\site\end{tabular} \\
\midrule
Plan-and-Act & 53.9 & 57.5 & 53.3 & 84.2 & \textbf{55.6} & 48.6 & 46.6 & 30 \\
\addlinespace
\begin{tabular}[l]{@{}l@{}}AgentOccam \textit{(Base)} \end{tabular} & 47.9 & 46.7 & 66.7 & 68.4 & 40 & 42.9 & 30.8 & 30 \\
\midrule
\textbf{Cognitive Map} \\
\midrule
\begin{tabular}[l]{@{}l@{}}  \textit{Base + CM-Raw} \end{tabular} & 44.8 & 47.1 & 70 & 68.4 & 35.6 & 51.4 & 19.2 & 0 \\
\addlinespace
\begin{tabular}[l]{@{}l@{}} \textit{Base + CM}\end{tabular} & 57.4 & 55.8 & 76.7 & 78.9 & 46.7 & 71.4 & 19.2 & 30 \\
\midrule
\textbf{Planning} \\
\midrule
\begin{tabular}[l]{@{}l@{}} \textit{Base + HL} \end{tabular} & 50.9 & 54.2 & 63.3 & 78.9 & 53.3 & 57.1 & 15.4 & 20 \\
\midrule
\textbf{WebATLAS} \\ \textit{Base+CM+HL+LA} & \textbf{63.0} & \textbf{67.1} & \textbf{73.3} & \textbf{84.2} & 53.3 & \textbf{77.1} & \textbf{42.3} & \textbf{40.0} \\
\bottomrule
\end{tabular}
% \line
% \addlinespace
\end{table}

%% file: 070conclusion.tex
\begin{comment}
\section{Conclusion}

This paper introduces a sophisticated web navigation agent that integrates explicit memory management with structured hierarchical planning, significantly enhancing exploration efficiency and decision-making effectiveness. By leveraging state-of-the-art Large Language Models, our approach provides robust contextual comprehension, supporting dynamic strategy adaptation across diverse web environments. Future work includes comprehensive empirical validation, further refinement of memory and planning architectures, and exploration of additional web navigation complexities and use-cases.
\end{comment}

\section{Conclusion}
This work presented \textbf{WebATLAS}, a web navigation agent that couples explicit, structured memory with hierarchical planning to turn open-ended browsing into a sequence of verifiable, low-entropy decisions. By leveraging contemporary large language models within a modular control loop, the agent maintains situational awareness across pages, decomposes goals into intermediate subgoals, and adapts its strategies as the interface or task constraints evolve. The result is a system that is not only more sample-efficient and time-efficient during exploration, but also more interpretable: intermediate memories, subplans, and decision rationales expose where and why the agent changes course.

\paragraph{Future Work} Looking forward, we hope to see a research agenda that emphasizes \emph{principled} generalization of this work rather than tuning performance on a single-benchmark. 

\begin{itemize}[leftmargin=*, labelsep=0.75em]
    \item First, our world-model representation of the web is still in its infancy. We hope to see others develop web-native world models that abstract repeated patterns (e.g., filters, tables, forms) into sub-programs and support counterfactual “what-if” reasoning, not merely retrieval. 
    \item Second, next-generation planning should be budget-aware and safety-aware by design, trading off success, latency, and risk through calibrated uncertainty and constraint handling. 
    \item Third, system robustness needs to be measured—not assumed—via stress tests that include UI drift, authentication flows, stochastic failures, and long-horizon, multi-session tasks. 
    \item Finally, as agentic systems start to close the gap with human performance, evaluation looking forward must move beyond pass/fail to incorporate cost of computation, side-effect penalties, reproducibility across seeds, and transparency of the intermediate state.
\end{itemize}

Taken together, these directions aim at agents that learn enduring abstractions of the web, plan under explicit budgets and constraints, and expose interpretable interfaces for verification and collaboration. We view this separation of concerns—memory, planning, and control—as a durable scaffold for the next generation of reliable, adaptable web agents which are certain to become ubiquitous and invaluable tools in the years to come.